\documentclass[10pt,twocolumn,letterpaper]{article}
\usepackage{iccv}
\usepackage{times}
\usepackage{epsfig}
\usepackage{graphicx}
\usepackage{amsmath}
\usepackage{amssymb}
\usepackage[utf8]{inputenc} 
\usepackage[T1]{fontenc}    
\usepackage{CJKutf8}        
\usepackage{booktabs}
\usepackage{url}            
\usepackage{booktabs}       
\usepackage{amsfonts}       
\usepackage{nicefrac}       
\usepackage{microtype}      
\usepackage{xcolor}
\usepackage{multirow}
\usepackage{tabularx}
\usepackage{adjustbox}
\usepackage{wrapfig}
\usepackage{caption}        

\usepackage[accsupp]{axessibility}

\usepackage{makecell}
\usepackage{pifont}
\usepackage[abs]{overpic}

\usepackage[pagebackref=true,breaklinks=true,letterpaper=true,colorlinks,bookmarks=false]{hyperref}

\usepackage{booktabs}

\usepackage[accsupp]{axessibility}

\iccvfinalcopy 

\def\iccvPaperID{2758} 

\begin{document}

\title{CLIP Guided Image-perceptive Prompt Learning for Image Enhancement}

\author{Weiwen Chen\textsuperscript{1} \qquad Qiuhong Ke\textsuperscript{2} \thanks{Corresponding authors} \qquad Zinuo Li\textsuperscript{2}
\\
Hongkong University\textsuperscript{1}, Monash University\textsuperscript{2} \\
{\tt\small Corresponding author: \href{Qiuhong.Ke@monash.edu}{Qiuhong.Ke@monash.edu}}
}

\hypersetup{
    colorlinks = true,
    linkbordercolor = {white},
    citecolor=blue
}

\newcommand{\lzn}[1]{{\color{blue}{[Zinuo: #1]}}}

\twocolumn[{%
   \renewcommand\twocolumn[1][]{#1}%
   \maketitle
   \vspace{-10mm}
   \begin{center}
    \centering
    \includegraphics[width=\linewidth]{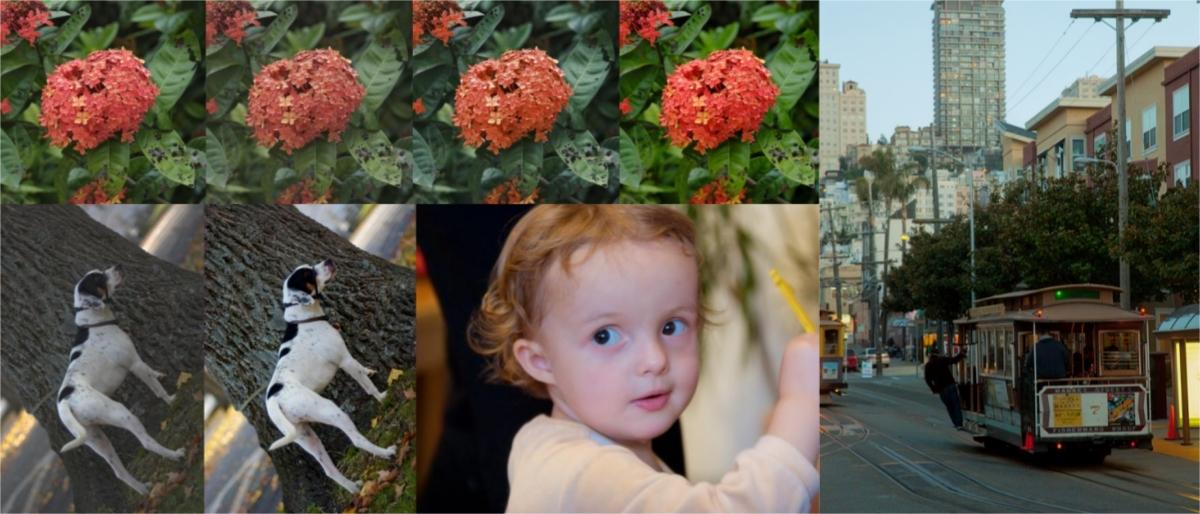}
    \vspace{-7mm}
    \captionof{figure}{The satisfactory visual results produced by the proposed CLIP-LUT. The four images in the top left corner come from the dataset FilmSet, the styles from left to right are Input, Cinema, ClassNeg and Velvia. The rest of the images are from the FiveK dataset.
 }
 \vspace{1mm}
    \label{fig:teaser}
   \end{center}%
}]

\ificcvfinal\thispagestyle{empty}\fi
\begin{abstract}
Image enhancement is a significant research area in the fields of computer vision and image processing. In recent years, many learning-based methods for image enhancement have been developed, where the Look-up-table (LUT) has proven to be an effective tool. In this paper, we delve into the potential of Contrastive Language-Image Pre-Training (CLIP) Guided Prompt Learning, proposing a simple structure called CLIP-LUT for image enhancement. We found that the prior knowledge of CLIP can effectively discern the quality of degraded images, which can provide reliable guidance. To be specific, We initially learn image-perceptive prompts to distinguish between original and target images using CLIP model, in the meanwhile, we introduce a very simple network by incorporating a simple baseline to predict the weights of three different LUT as enhancement network. The obtained prompts are used to steer the enhancement network like a loss function and improve the performance of model. We demonstrate that by simply combining a straightforward method with CLIP, we can obtain satisfactory results.

\end{abstract}

\section{Introduction}
Image enhancement plays a pivotal role in the field of digital image processing and computer vision. Its primary goal is to improve the quality and interpretability of images, thereby enabling better human perception and facilitating more accurate machine processing. Techniques for image enhancement can range from basic adjustments of contrast and colour balance to more complex operations that improve the visibility of details in an image or introduce desirable transformations.

In recent years, learning-based methods have emerged as powerful tools for image enhancement, offering the potential to learn complex transformations from data and thus achieve superior results. These methods often leverage deep neural networks and have been applied to a variety of tasks~\cite{chen2018deep}~\cite{wang2019underexposed}~\cite{zhang2021star}~\cite{gharbi2017deep}. 3DLUT~\cite{zeng2020learning} and SepLUT~\cite{yang2022seplut} are based on the Look-Up-Table (LUT), a reliable tool for automatic image colour grading. LPTN~\cite{liang2021high} and DeepLPF~\cite{moran2020deeplpf} respectively enhance images by converting low-frequency elements with reduced resolution and using learning space local filters.

Despite these advancements, the potential of visual-language models, specifically Contrastive Language–Image Pretraining (CLIP), in image enhancement remains largely untapped. These models, which exhibit an integrated understanding of visual and textual data, have the potential to provide context-rich features and guidance for enhancement tasks, yet there has been relatively little exploration in this area.

In this study, we address this gap by delving into the potential of Contrastive Language-Image Pre-Training (CLIP) Guided Prompt Learning, proposing CLIP-LUT for low-level image enhancement. We found that CLIP is particularly suitable for distinguishing the quality of synthesized dataset such as FilmSet~\cite{li2023large}. Our approach utilizes CLIP to discern image quality and get image-perceptive prompts, together with a straightforward a simple baseline to predict LUT weights, treating learned prompts as a guiding loss function. By simply combining CLIP and a simple method, we achieved decent performance. We believe this approach could open up new possibilities for image enhancement and further understanding of CLIP in low-level computer vision.

\section{Related Work}

\noindent\textbf{Image Enhancement}
In recent years, many exciting image enhancement methods have appeared. DPE model~\cite{chen2018deep} utilizes a U-Net model to facilitate the learning of pixel-to-pixel correspondences between input and output image pairs. Conversely, UPE model~\cite{wang2019underexposed} attempts to estimate the pixel-wise illumination map derived from the input image. HDRNet~\cite{gharbi2017deep} is designed to determine pixelwise transformation coefficients within the bilateral space. 3D LUTs~\cite{zeng2020learning} have found extensive application in adjusting the colour and tone of photographic images. SepLUT~\cite{yang2022seplut} dissects a solitary colour transformation into two distinct sub-transformations: those that are independent of components and those that are correlated with components, with the aim to augment image quality. Based on the Laplacian Pyramid, LPTN~\cite{liang2021high} is devised to transpose the low-frequency elements of diminished resolution while concurrently honing the high-frequency constituents with efficiency. DeepLPF~\cite{moran2020deeplpf} employs three different types of learning space local filters to automatically enhance images. STAR-DCE~\cite{zhang2021star} is mainly used for image and video enhancement on smartphones. Despite the aforementioned methods, the exploration of CLIP in low-level enhancement still remains a relatively uncharted area.

\noindent\textbf{CLIP and Visual Prompting}
CLIP model~\cite{radford2021learning} in the realm of zero-shot classification are notable, largely attributable to its extensive training on a diverse array of image-text data. Its adaptability and proficiency have been effectively demonstrated across various high-level tasks ~\cite{zang2022open}~\cite{kuo2022f}~\cite{zhou2022extract}. In a notable exploration~\cite{wang2023exploring}, it was discovered that the comprehensive visual language priors inherent in CLIP could be harnessed for the dual purpose of appraising both image quality and abstract perception in a zero-shot context. These insights lay the foundation for our work, which seeks to utilize CLIP for the specific task of enhancing backlit images. VPT~\cite{jia2022visual} is a method characterized by efficient parameter utilization, enabling the use of expansive visual Transformer models for a range of downstream tasks. By integrating task-specific learnable prompts, VPT outperforms competing fine-tuning methods, achieving superior performance while also curtailing storage expenditure. In the landscape of vision-and-language models, prompt learning has emerged as an intriguing area of research with considerable potential. This was first exemplified by the CoOp model~\cite{zhou2022learning}, which incorporated prompt learning into the adaptation process of vision-language models for downstream vision tasks. The CoCoOp model~\cite{zhou2022conditional} built on this, innovating by conditioning the prompt on individual input instances, thus enhancing the model's flexibility in contrast to static prompts. Yet, the focus of existing methodologies remains on high-level vision tasks. In this paper, in contrast, we leverage prompt learning to distill refine image. This approach aims to propel the application of CLIP guided prompt learning methodologies into the domain of low-level image enhancement, further enriching this dynamic field of research.

\section{Methodology}
\begin{figure*}[ht]
    \begin{minipage}[b]{1.0\linewidth}
        \includegraphics[width=\linewidth]{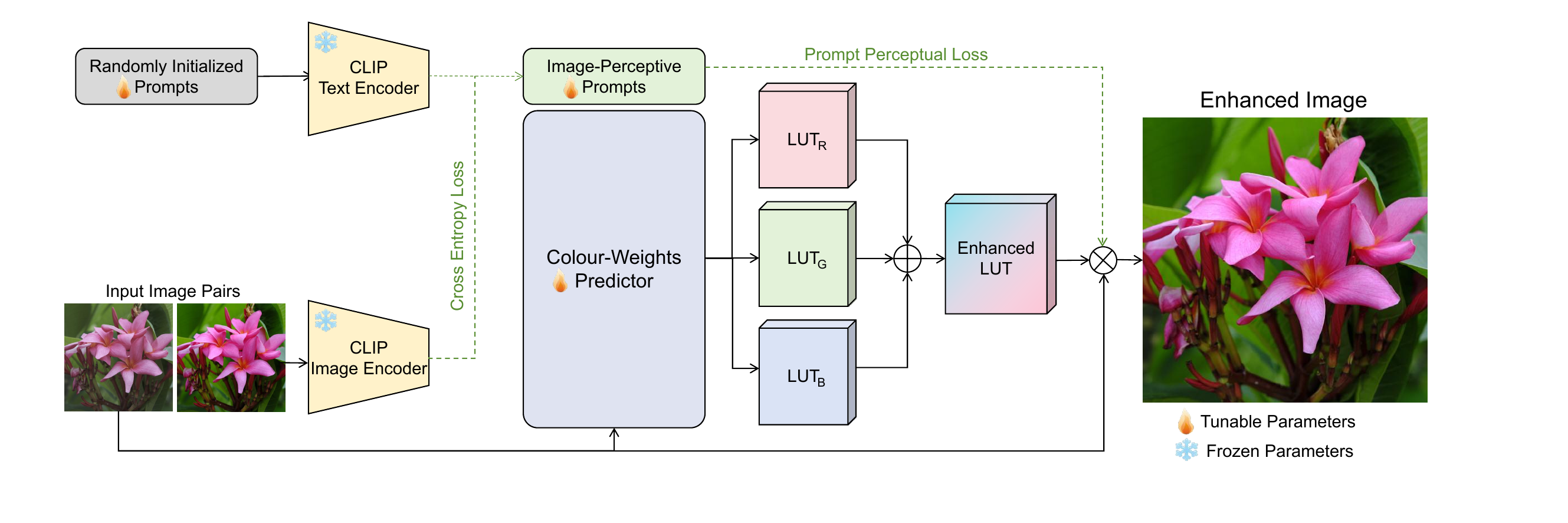}
    \end{minipage}
    \vspace{-1em}
    \caption{
    The simple network structure of CLIP-LUT. We first randomly initialize prompts and send them along with the input image pairs to the frozen CLIP Text Encoder and Image Encoder for training, this process is controlled by cross-entropy loss. In this stage, we learn image-perceptive prompts that can accurately distinguish the original images and enhanced images. These prompts will then guide the learning of the network as a loss function to predict the weights of the three LUTs. Finally, multiplying the input image with the enhanced LUT yields the final result.
    }
    \label{fig:method}
\end{figure*}

\noindent\textbf{Pipline Overview}
The overall pipline is shown in Figure~\ref{fig:method}. The whole pipeline is divided into two parts: prompt learning and image enhancement. In the prompt learning stage, we first randomly initialize a set of prompts from a normal distribution and input them to the frozen CLIP Text Encoder to obtain their latent encodings in CLIP space. At the same time, we input the image pairs (original images and enhanced images) to the frozen CLIP Image Encoder to obtain their latent image encoding. The prompts are trained to efficiently distinguish the original images and enhanced images, we refer to the prompts in this stage as image-perceptive prompts. In the subsequent stages, these prompts will serve as guidance for learning as part of the loss function. In the image enhancement stage, we use a simple baseline as a colour-weight predictor to learn the weights of three different LUTs, which are then combined into one LUT to directly refine the image.

\noindent\textbf{Prompt Learning}
Inspired by~\cite{liang2023iterative}, which has proven the efficiency of prompt learning in low-level computer vision, we also use frozen CLIP for prompt learning in this stage. Given an image pair of original image $I_{o} \in \mathbb{R}^{H\times W \times 3}$ and enhanced image $I_{e} \in \mathbb{R}^{H\times W \times 3}$, we randomly initialize the original prompt $P_{o} \in \mathbb{R}^{N\times 512}$ and the enhancement prompt $P_{e} \in \mathbb{R}^{N\times 512}$ using normal distribution. In this stage, the target prompts for enhanced images are set to 1, while the target prompts for the original images are set to 0. Under the control of loss function, the learning aim is to make the prompts give scores as close to 1 as possible when facing enhanced images, and scores as close to 0 as possible when facing original images. 

We utilize Cross Entropy Loss in CLIP latent space to get the similarity between text encoding and image encoding for learning the image-perceptive prompts, which can be described as Equation~\ref{eq:cross_entropy} and Equation~\ref{eq:label}: 
\begin{equation}
    \mathcal{L}_{prompts}= -(y*\log(\hat{y})+(1-y)*\log(1-\hat{y}))
    \label{eq:cross_entropy}
\end{equation}
\begin{equation}
    \hat{y}=\frac{e^{cos(\Phi_{image}(I),\Phi_{text}(T_e))}}{\sum_{i\in{\{o,e\}}}e^{cos(\Phi_{image}(I),\Phi_{text}(T_i))}}
    \label{eq:label}
\end{equation}
where $I \in \{I_{o}, I_{e}\}$, $y$ means the label . Here $y = 0$ is for negative sample $I_{o}$ and $y = 1$ is for positive sample $I_{e}$.

\noindent\textbf{Image Enhancement}
In this stage, we utilize a simple UNet-like baseline~\cite{chen2022simple} as the colour-weights predictor. The colour-weights predictor epitomizes a streamlined non-linear network characterized by its lightweight parameters, effectively obviating the necessity for superfluous activation functions including but not limited to Sigmoid, Softmax, and ReLU. Empirical evidence has substantiated that this modification does not precipitate a decline in performance. This is succeeded by a series of two convolutions. Post the implementation of deformable convolution, elements such as SimpleGate and the Simplified Channel Attention (SCA) are deployed to enhance the performance.

\begin{table*}[!ht]
\centering
\resizebox{0.9\linewidth}{!}{
\begin{tabular}{c|ccc|ccc|ccc}
\hline
                         & \multicolumn{3}{c|}{Cinema}                                                                                                         & \multicolumn{3}{c|}{ClassNeg}                                                                                                       & \multicolumn{3}{c}{Velvia}                                                                                                          \\ \cline{2-10} 
\multirow{-2}{*}{Method} & \multicolumn{1}{c|}{PSNR$\uparrow$}               & \multicolumn{1}{c|}{SSIM$\uparrow$}               & $\Delta E$$\downarrow$      & \multicolumn{1}{c|}{PSNR$\uparrow$}               & \multicolumn{1}{c|}{SSIM$\uparrow$}               & $\Delta E$$\downarrow$      & \multicolumn{1}{c|}{PSNR$\uparrow$}               & \multicolumn{1}{c|}{SSIM$\uparrow$}               & $\Delta E$$\downarrow$      \\ \hline
HDRNet                   & \multicolumn{1}{c|}{35.18}                        & \multicolumn{1}{c|}{0.990}                        & 2.81                        & \multicolumn{1}{c|}{35.41}                        & \multicolumn{1}{c|}{0.988}                        & 2.19                        & \multicolumn{1}{c|}{34.37}                        & \multicolumn{1}{c|}{0.975}                        & 3.56                        \\
DPE                      & \multicolumn{1}{c|}{3.98}                         & \multicolumn{1}{c|}{0.358}                        & 47.58                       & \multicolumn{1}{c|}{3.79}                         & \multicolumn{1}{c|}{0.320}                        & 49.66                       & \multicolumn{1}{c|}{3.48}                         & \multicolumn{1}{c|}{0.313}                        & 52.12                       \\
UPE                      & \multicolumn{1}{c|}{22.81}                        & \multicolumn{1}{c|}{0.946}                        & 4.22                        & \multicolumn{1}{c|}{22.50}                        & \multicolumn{1}{c|}{0.936}                        & 4.97                        & \multicolumn{1}{c|}{22.23}                        & \multicolumn{1}{c|}{0.893}                        & 5.00                        \\
DeepLPF                  & \multicolumn{1}{c|}{36.34}                        & \multicolumn{1}{c|}{0.985}                        & 1.96                        & \multicolumn{1}{c|}{33.40}                        & \multicolumn{1}{c|}{0.978}                        & 2.43                        & \multicolumn{1}{c|}{34.06}                        & \multicolumn{1}{c|}{0.956}                        & 2.24                        \\
3D-LUT                   & \multicolumn{1}{c|}{35.49}                        & \multicolumn{1}{c|}{0.990}                        & 1.86                        & \multicolumn{1}{c|}{33.82}                        & \multicolumn{1}{c|}{0.989}                        & 1.83                        & \multicolumn{1}{c|}{34.07}                        & \multicolumn{1}{c|}{0.976}                        & 2.40                        \\
STAR-DCE                 & \multicolumn{1}{c|}{28.12}                        & \multicolumn{1}{c|}{0.949}                        & 6.91                        & \multicolumn{1}{c|}{25.54}                        & \multicolumn{1}{c|}{0.945}                        & 7.98                        & \multicolumn{1}{c|}{34.06}                        & \multicolumn{1}{c|}{0.956}                        & 2.24                        \\
LPTN                   & \multicolumn{1}{c|}{36.55}                        & \multicolumn{1}{c|}{0.985}                        & 2.12                        & \multicolumn{1}{c|}{34.22}                        & \multicolumn{1}{c|}{0.972}                        & 2.72                        & \multicolumn{1}{c|}{33.19}                        & \multicolumn{1}{c|}{0.948}                        & 3.32                        \\
SepLUT                   & \multicolumn{1}{c|}{35.82}                        & \multicolumn{1}{c|}{0.986}                        & 2.42                        & \multicolumn{1}{c|}{34.10}                        & \multicolumn{1}{c|}{0.982}                        & 2.34                        & \multicolumn{1}{c|}{32.88}                        & \multicolumn{1}{c|}{0.964}                        & 3.60                       \\
FilmNet                     & \multicolumn{1}{c|}{{\color[HTML]{CB0000} \textbf{40.07}}} & \multicolumn{1}{c|}{{0.993}} & {1.61} & \multicolumn{1}{c|}{{38.89}} & \multicolumn{1}{c|}{{0.992}} & {1.47} & \multicolumn{1}{c|}{{37.60}} & \multicolumn{1}{c|}{{0.981}} & {2.05} \\
CLIP-LUT (Ours)                     & \multicolumn{1}{c|}{{\underline{39.85}}} & \multicolumn{1}{c|}{{\color[HTML]{CB0000} \textbf{0.994}}} & {\color[HTML]{CB0000} \textbf{1.48}} & \multicolumn{1}{c|}{{\color[HTML]{CB0000} \textbf{39.05}}} & \multicolumn{1}{c|}{{\color[HTML]{CB0000} \textbf{0.994}}} & {\color[HTML]{CB0000} \textbf{1.44}} & \multicolumn{1}{c|}{{\color[HTML]{CB0000} \textbf{37.68}}} & \multicolumn{1}{c|}{{\color[HTML]{CB0000} \textbf{0.982}}} & {\color[HTML]{CB0000} \textbf{1.95}} \\ \hline
\end{tabular}
}
\caption{Quantitative comparisons on the FilmSet dataset of different image enhancement methods. The top result is highlighted in red and second is underlined.}
\label{table:exp_2}
\end{table*}

\begin{table*}[!ht]
\centering
\resizebox{0.9\linewidth}{!}{
\begin{tabular}{c|ccc|ccc|ccc}
\hline
                         & \multicolumn{3}{c|}{Cinema}                                                                                                         & \multicolumn{3}{c|}{ClassNeg}                                                                                                       & \multicolumn{3}{c}{Velvia}                                                                                                          \\ \cline{2-10} 
\multirow{-2}{*}{Method} & \multicolumn{1}{c|}{PSNR$\uparrow$}               & \multicolumn{1}{c|}{SSIM$\uparrow$}               & $\Delta E$$\downarrow$      & \multicolumn{1}{c|}{PSNR$\uparrow$}               & \multicolumn{1}{c|}{SSIM$\uparrow$}               & $\Delta E$$\downarrow$      & \multicolumn{1}{c|}{PSNR$\uparrow$}               & \multicolumn{1}{c|}{SSIM$\uparrow$}               & $\Delta E$$\downarrow$      \\ 
\hline
W/O Prompts                 & \multicolumn{1}{c|}{37.68}                        & \multicolumn{1}{c|}{0.987}                        & 2.14                        & \multicolumn{1}{c|}{35.40}                        & \multicolumn{1}{c|}{0.981}                        & 2.26                        & \multicolumn{1}{c|}{36.54}                        & \multicolumn{1}{c|}{0.971}                        & 2.17                        \\
Random Prompts              & \multicolumn{1}{c|}{31.46}                        & \multicolumn{1}{c|}{0.932}                        & 2.89                        & \multicolumn{1}{c|}{29.65}                        & \multicolumn{1}{c|}{0.907}                        & 3.16                        & \multicolumn{1}{c|}{30.60}                        & \multicolumn{1}{c|}{0.913}                        & 2.97                        \\
Ours                     & \multicolumn{1}{c|}{{\color[HTML]{CB0000} \textbf{39.85}}} & \multicolumn{1}{c|}{{\color[HTML]{CB0000} \textbf{0.994}}} & {\color[HTML]{CB0000} \textbf{1.48}} & \multicolumn{1}{c|}{{\color[HTML]{CB0000} \textbf{39.05}}} & \multicolumn{1}{c|}{{\color[HTML]{CB0000} \textbf{0.994}}} & {\color[HTML]{CB0000} \textbf{1.44}} & \multicolumn{1}{c|}{{\color[HTML]{CB0000} \textbf{37.68}}} & \multicolumn{1}{c|}{{\color[HTML]{CB0000} \textbf{0.982}}} & {\color[HTML]{CB0000} \textbf{1.95}} \\ \hline
\end{tabular}
}
\caption{Ablation studies on the FilmSet dataset of different image enhancement methods. The top result is highlighted in red.}
\label{table:exp_2ab}
\end{table*}

SimpleGate seperates the features directly into dual components along the channel dimension, subsequently effectuating a multiplication operation on them. Meanwhile, the Simplified Channel Attention (SCA) harnesses the straightforward $1 \times 1$ convolution methodology for channel-wise data transmission. The functional paradigm of SimpleGate can be articulated through the subsequent Equation~\ref{eq:sg} and Equation~\ref{eq:sca}:
\begin{equation}
\operatorname{SimpleGate}(\mathbf{X}, \mathbf{Y})=\mathbf{X} \odot \mathbf{Y} , 
\label{eq:sg}
\end{equation}
\begin{equation}
SCA(\mathbf{X})=\mathbf{X} * W \operatorname{pool}(\mathbf{X})
\label{eq:sca}
\end{equation}
where X and Y are identically sized feature maps, $\odot$ is an element-wise multiplication, $W$ is fully-connected layer, $pool$ represents the global average pooling procedure that combines spatial data into channels, $*$ indicates a channel-wise multiplication.

Given a 3-channel source image with RGB colour $\left\{r_{(x, y, z)}^{I}, g_{(x, y, z)}^{I}, b_{(x, y, z)}^{I}\right\}$, a LUT is executed to ascertain the respective coordinates $(x, y, z)$ within the 3-dimensional LUT lattice framework, as delineated in the Equation~\ref{eq:lut}:
\begin{equation}
x=\frac{r_{(x, y, z)}^{I}}{s}, y=\frac{g_{(x, y, z)}^{I}}{s}, z=\frac{b_{(x, y, z)}^{I}}{s}
\label{eq:lut}
\end{equation}
where $s =\frac{C_{max}}{M}$, $C_{max}$ indicates the maximum colour value and $M$ is the number of bins in each colour channel. 

As shown in Figure~\ref{fig:method}, the colour-weights predictor predicts the weights of three different LUTs represent RGB channels following the Equation~\ref{eq:elut}:
\begin{equation}
    \mathbf{LUT}_{e}= \omega_1 * \mathbf{LUT_R} + \omega_2 * \mathbf{LUT_G} + \omega_3 * \mathbf{LUT_B}
    \label{eq:elut}
\end{equation}
where $\omega_1$, $\omega_2$ and $\omega_3$ are the weights of three LUTs respectively, $\mathbf{LUT}_{e}$ represents the Enhanced LUT. The final enhanced image is obtained by multiplying the original image by the $\mathbf{LUT}_{e}$.

Finally, we incorporate prompt perceptual loss as Equation~\ref{eq:per}. In the whole training phase, we combine $\mathcal{L}_{MSE}$, $\mathcal{L}_{SSIM}$ and $\mathcal{L}_{perceptual}$ into a total loss function as Equation~\ref{eq:total}:
\begin{equation}
    \mathcal{L}_{perceptual}=\frac{e^{cos(\Phi_{image}(I),\Phi_{text}(T_o))}}{\sum_{i\in{\{o,e\}}}e^{cos(\Phi_{image}(I),\Phi_{text}(T_i))}}
    \label{eq:per}
\end{equation}
\begin{equation}
\mathcal{L}_{total} = \mathcal{L}_{MSE} + 0.4 * \mathcal{L}_{perceptual} + 0.4 * \mathcal{L}_{SSIM}
\label{eq:total}
\end{equation}
To balance the composition of $\mathcal{L}_{total}$, we set the weight of $\mathcal{L}_{perceptual}$ to 0.4, as it is too large compared to $\mathcal{L}_{MSE}$ and $\mathcal{L}_{SSIM}$. In this case, the value ranges of $\mathcal{L}_{MSE}$ and $\mathcal{L}_{perceptual}$ are roughly similar, thus the loss function is jointly dominated by $\mathcal{L}_{MSE}$ and $\mathcal{L}_{perceptual}$, with $\mathcal{L}_{SSIM}$ playing a auxiliary role.

\begin{figure*}[!ht]
    \begin{minipage}[b]{1.0\linewidth}
        \begin{minipage}[b]{0.24\linewidth}
            \centering
            \centerline{\includegraphics[width=4.45cm]{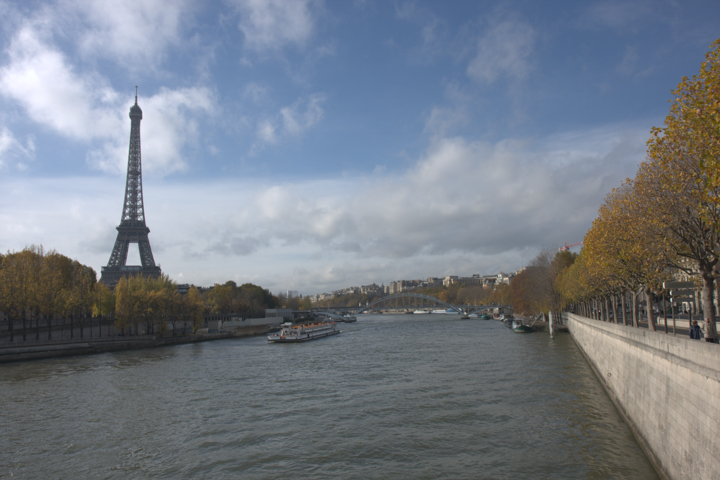}}
            \centerline{(a) Input}\medskip
        \end{minipage}
        \hfill
        \begin{minipage}[b]{0.24\linewidth}
            \centering
            \centerline{\includegraphics[width=4.45cm]{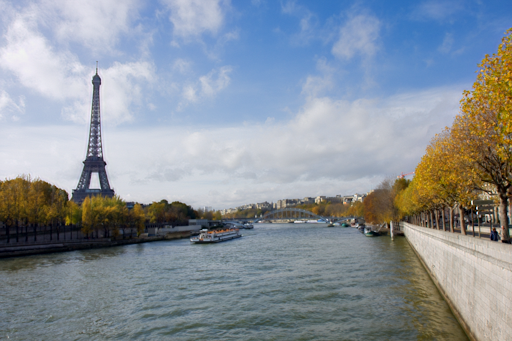}}
            \centerline{(b) DPE}\medskip
        \end{minipage}
        \hfill
        \begin{minipage}[b]{0.24\linewidth}
            \centering
            \centerline{\includegraphics[width=4.45cm]{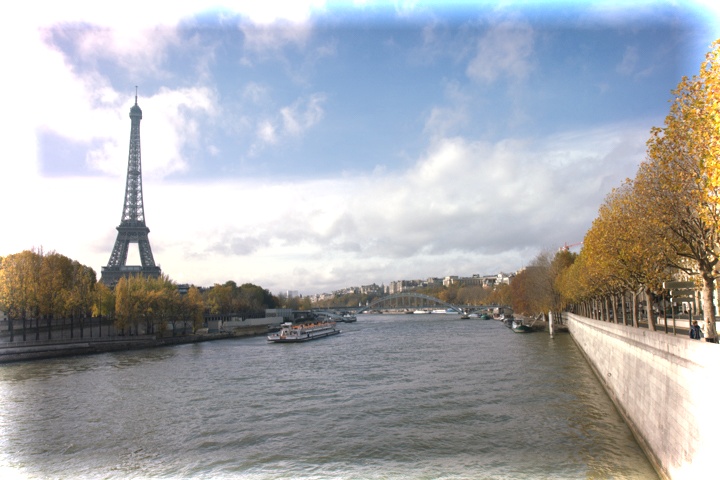}}
            \centerline{(c) UPE}\medskip
        \end{minipage}
        \hfill
        \begin{minipage}[b]{0.24\linewidth}
            \centering
            \centerline{\includegraphics[width=4.45cm]{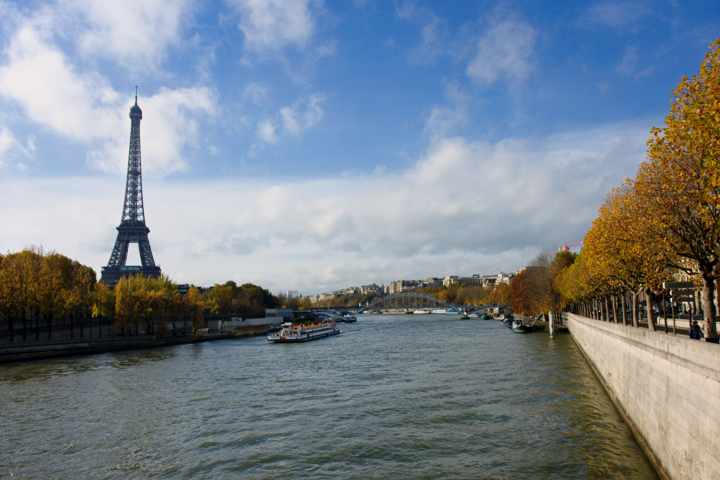}}
            \centerline{(d) LPTN}\medskip
        \end{minipage}
    \end{minipage} 

    \begin{minipage}[b]{1.0\linewidth}
        \begin{minipage}[b]{0.24\linewidth}
            \centering
            \centerline{\includegraphics[width=4.45cm]{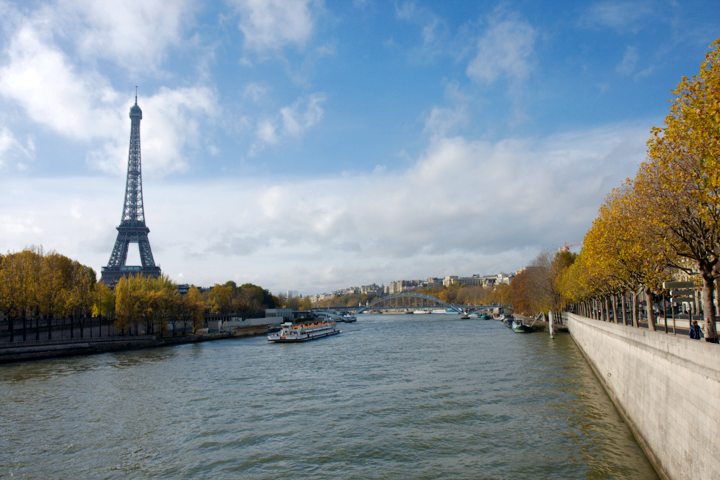}}
            \centerline{(e) 3D-LUT}\medskip
        \end{minipage}
        \hfill
        \begin{minipage}[b]{0.24\linewidth}
            \centering
            \centerline{\includegraphics[width=4.45cm]{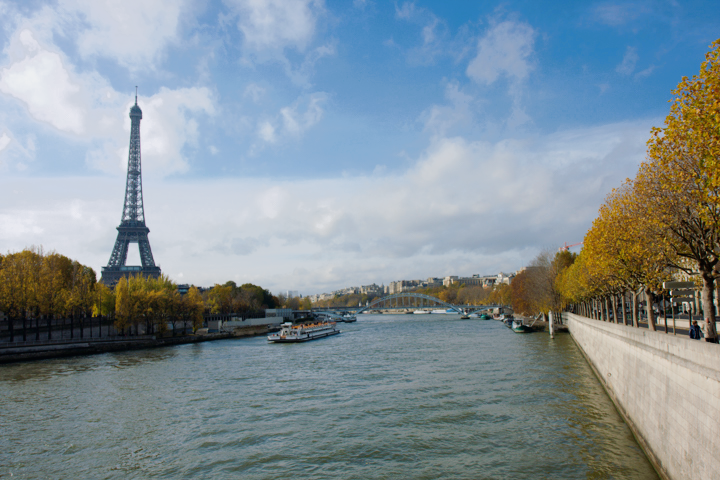}}
            \centerline{(f) SepLUT}\medskip
        \end{minipage}
        \hfill
        \begin{minipage}[b]{0.24\linewidth}
            \centering
            \centerline{\includegraphics[width=4.45cm]{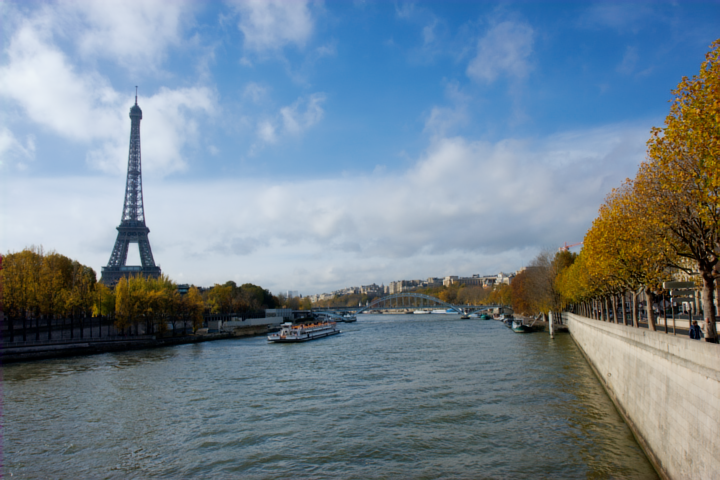}}
            \centerline{(g) Ours}\medskip
        \end{minipage}
        \hfill
        \begin{minipage}[b]{0.24\linewidth}
            \centering
            \centerline{\includegraphics[width=4.45cm]{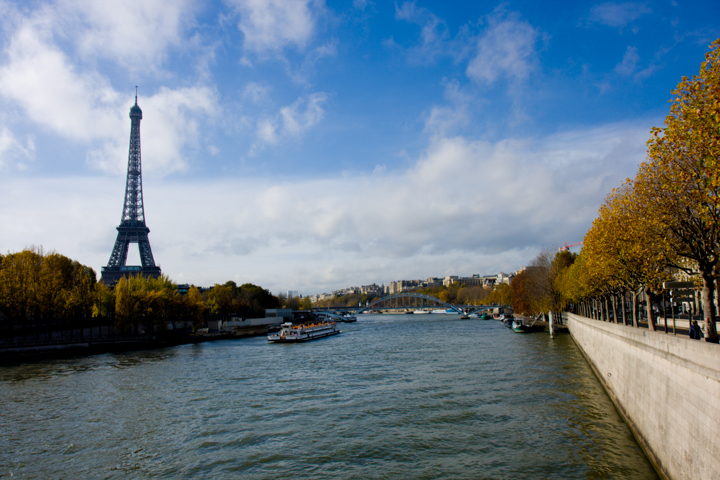}}
            \centerline{(h) Target}\medskip
        \end{minipage}
    \end{minipage} 
    
    \caption{
    Visual comparison of various image enhancement methods on the FiveK dataset. Our result~(g) is aesthetically superior in terms of color tone and specifics. The DPE~(b) and UPE~(c) results greatly deviate from the objective. Their color, exposure and reproduction of fine details are unsatisfactory. 3D-LUT~(e), SepLUT~(f) and LPTN~(d) are visually better, however the tone mapping is typically too bright or too dark compared to the target~(h).
    }
    \label{fig:fivek}
\end{figure*}
\begin{table*}[!ht]
\centering
\resizebox{0.8\linewidth}{!}{
\begin{tabular}{c|ccc|ccc}
\hline
                         & \multicolumn{3}{c|}{Fivek}                                                                                                     & \multicolumn{3}{c}{HDR+}                                                                                \\ \cline{2-7} 
\multirow{-2}{*}{Method} & \multicolumn{1}{c|}{{\color[HTML]{000000} PSNR$\uparrow$}} & \multicolumn{1}{c|}{SSIM$\uparrow$} & $\Delta E$$\downarrow$      & \multicolumn{1}{c|}{PSNR$\uparrow$} & \multicolumn{1}{c|}{SSIM$\uparrow$} & $\Delta E$$\downarrow$      \\ \hline
HDRNet                   & 19.93                                                      & 0.798                               & 14.42                       & 23.04                               & 0.879                               & 8.97                        \\
DPE                      & 17.66                                                      & 0.725                               & 17.71                       & 22.56                               & 0.872                               & 10.45                       \\
UPE                      & 21.88                                                      & 0.853                               & 10.80                       & 21.21                               & 0.816                               & 13.05                       \\
DeepLPF                  & 24.55                                                      & 0.846                               & 8.62                        & N/A                                 & N/A                                 & N/A                         \\
3D-LUT                   & 24.59                                                      & 0.846                               & 8.30                        & 23.54                               & 0.885                               & 7.93                        \\
STAR-DCE                 & 24.50                                                       & 0.893                               & N/A                         & {\color[HTML]{CB0000} \textbf{26.50}}                                & 0.883                               & 5.77                        \\
LPTN                   & 22.19                                                      &  0.878                               & 11.90                        & N/A                                 & N/A                                 & N/A                          \\
SepLUT                   & 25.02                                                      & 0.873                               & 7.91                        & N/A                                 & N/A                                 & N/A                         \\
Ours                     & {\color[HTML]{CB0000} \textbf{25.34}}                               & {\color[HTML]{CB0000} \textbf{0.922}}        & {\color[HTML]{CB0000} \textbf{7.26}} & {\underline{26.39}}        & {\color[HTML]{CB0000} \textbf{0.891}}        & {\color[HTML]{CB0000} \textbf{5.56}} \\ \hline
\end{tabular}
}
\caption{Quantitative comparisons on the MIT FiveK and HDR+ dataset of different image enhancement methods. "N/A" indicates that the result is unavailable and the top result is highlighted in red and second is underlined.}
\label{table:exp_1}
\end{table*}
\begin{table*}[!ht]
\centering
\resizebox{0.8\linewidth}{!}{
\begin{tabular}{c|ccc|ccc}
\hline
                         & \multicolumn{3}{c|}{Fivek}                                                                                                     & \multicolumn{3}{c}{HDR+}                                                                                \\ \cline{2-7} 
\multirow{-2}{*}{Method} & \multicolumn{1}{c|}{{\color[HTML]{000000} PSNR$\uparrow$}} & \multicolumn{1}{c|}{SSIM$\uparrow$} & $\Delta E$$\downarrow$      & \multicolumn{1}{c|}{PSNR$\uparrow$} & \multicolumn{1}{c|}{SSIM$\uparrow$} & $\Delta E$$\downarrow$      \\ 
                        \hline
W/O Prompts                 & 24.37                                                      & 0.873                               & 8.71                        & 23.97                               & 0.892                               & 6.71                        \\
Random Prompts               & 21.14                                                      & 0.813                                & 9.63                       & 19.08                               & 0.786                               & 10.24                         \\
Ours                     & {\color[HTML]{CB0000} \textbf{25.34}}                               & {\color[HTML]{CB0000} \textbf{0.922}}        & {\color[HTML]{CB0000} \textbf{7.26}} & {\color[HTML]{CB0000} \textbf{26.39}}        & {\color[HTML]{CB0000} \textbf{0.891}}        & {\color[HTML]{CB0000} \textbf{5.56}} \\ \hline
\end{tabular}
}
\caption{Ablation studies on the MIT FiveK and HDR+ dataset of different image enhancement methods. The top result is highlighted in red.}
\label{table:exp_1ab}
\end{table*}

\section{Expriments}

\subsection{Experimental Setup}
\begin{table}[!ht]
\centering
\begin{tabular}{c|c|c|c}
\hline
Dataset  & FiveK & HDR+ & FilmSet \\ \hline
Training & 4502  & 675  & 4657    \\ \hline
Testing  & 498   & 247  & 638     \\ \hline
\end{tabular}
\caption{The configuration of three datasets.}
\label{table:dataset}
\end{table}
In this section, we utilize three distinct datasets for both training and evaluation purposes, namely the MIT-Adobe FiveK~\cite{bychkovsky2011learning}, HDR+\cite{hasinoff2016burst} and FilmSet~\cite{li2023large}. The MIT-Adobe FiveK dataset stands as the most extensive resource available for image enhancement, encompassing five retouched renditions of 5,000 original images under diverse conditions. The HDR+ dataset, originating from the Google Camera Group, provides 3,640 scenes designed for high dynamic range and low-light enhancement, specifically tailored for burst photography. FilmSet is a comprehensive, high-quality dataset that features over 8,000 images spanning three unique film genres. The dataset configuration is shown in Table~\ref{table:dataset}:

The implementation is base on only single A40 with image batchsize of 8 and prompt batchsize of 16. We first train image-percepetive prompts with 100 epochs and then enhancment network is trained with 200 epochs. To accelerate the training and validation processes, all images have been reformatted to a $256 \times 256$ resolution and the standard PNG format. 

We assess frameworks by employing the Peak Signal-to-Noise Ratio (PSNR), Structural Similarity (SSIM), and $\Delta E$ metrics. Notably, $\Delta E$ quantifies colour discrepancy as perceived by humans within the CIELab colour space~\cite{backhaus2011color}. Elevated PSNR and SSIM values are indicative of amplified performance, while a diminished $\Delta E$ suggests improved colour representation.

\subsection{Comparative Experiments}
In this section, we incorporate nine methodologies: HDRNet~\cite{gharbi2017deep}, DPE~\cite{chen2018deep}, UPE~\cite{wang2019underexposed}, DeepLPF~\cite{moran2020deeplpf}, 3D-LUT~\cite{zeng2020learning}, STAR-DCE~\cite{zhang2021star}, LPTN~\cite{liang2021high}, SepLUT~\cite{yang2022seplut} and FilmNet~\cite{li2023large}. For the methods that provide pretrained models with weights files on corresponding datasets, we test using their pretrained models; otherwise we retrain their official code on the dataset.


\subsection{Ablation Studies}
In this section, we conducted three different ablation experiments. \textbf{W/O Prompts} represents training without prompts guidance. \textbf{Random Prompts} means training with a randomly generated prompt from a normal distribution. \textbf{Ours} represents our proposed CLIP-LUT method, which uses image-perceptive prompts to guide the network.

\section{Limitation}
Although we have demonstrated in this experiment that using CLIP to guide training with prompts can effectively guide network training and generate satisfactory results, the idea presented in this paper is still in an exploratory stage, and more effective methods remain to be explored. For example, there may be better ways to learn prompts, the loss functions used to guide learning with prompts could have better choices, and the networks used for enhancement under could become more lightweight.

\section{Conclusion}
In this work, we explore the capabilities of CLIP Guided Prompt Learning, introducing a straightforward structure termed as CLIP-LUT for image enhancement. Our findings suggest that the intrinsic knowledge of CLIP can effectively discern the quality of deteriorated images, thereby providing reliable guidance. We initially learn prompts to differentiate between original and target images using the CLIP model. In the meanwhile, we incorporate a simple baseline to predict the weights of three distinct LUTs as an enhancement network. These acquired prompts are utilized to guide the enhancement network akin to a loss function, thereby enhancing the model's performance. We illustrate that the simple amalgamation of a straightforward methodology with CLIP can yield satisfactory results. We believe incorporating CLIP in low-level computer vision can be more efficiently and effectively in the future.

{\small
\bibliographystyle{ieee_fullname}
\bibliography{egbib}
}

\end{document}